\pdfoutput=1

\documentclass[11pt]{article}

\usepackage[final]{acl}

\usepackage{times}
\usepackage{latexsym}

\usepackage[T1]{fontenc}

\usepackage[utf8]{inputenc}

\usepackage{microtype}

\usepackage{inconsolata}

\usepackage{graphicx}

\usepackage{tikz}
\usepackage{pgfplots}
\usepackage{pgfplotstable}
\usepackage{listings}
\usepackage{xcolor} 
\usepackage{csquotes}

\usetikzlibrary{patterns}

\title{A Retail-Corpus for Aspect-Based Sentiment Analysis with Large Language Models}

\author{
  \textbf{Oleg Şilcenco\textsuperscript{1}},
  \textbf{Marcos R. Machado\textsuperscript{1}},
  \textbf{Wallace C. Ugulino\textsuperscript{1}},
  \textbf{Daniel Braun\textsuperscript{2}}
\\
 \textsuperscript{1}University of Twente,
 \textsuperscript{2}Marburg University
 \\
  \small{
    olegsilcenco@gmail.com, \{m.r.machado,w.corbougulino\}@utwente.nl, daniel.braun@uni-marburg.de
  }
}

\makeatletter
\pgfplotsset{
    /pgfplots/flexible yticklabels from table/.code n args={3}{%
        \pgfplotstableread[#3]{#1}\coordinate@table
        \pgfplotstablegetcolumn{#2}\of{\coordinate@table}\to\pgfplots@yticklabels
        \let\pgfplots@yticklabel=\pgfplots@user@ticklabel@list@y
    }
}
\makeatother

\begin{document}

\maketitle
\begin{abstract}
Aspect-based sentiment analysis enhances sentiment detection by associating it with specific aspects, offering deeper insights than traditional sentiment analysis. This study introduces a manually annotated dataset of 10,814 multilingual customer reviews covering brick-and-mortar retail stores, labeled with eight aspect categories and their sentiment. Using this dataset, the performance of GPT-4 and LLaMA-3 in aspect based sentiment analysis is evaluated to establish a baseline for the newly introduced data. The results show both models achieving over 85\% accuracy, while GPT-4 outperforms LLaMA-3 overall with regard to all relevant metrics.
\end{abstract}

\section{Introduction}

Sentiment analysis, i.e. the automatic identification of the sentiment expressed in, e.g., a text, is a widely used technique in research, business, politics, and many other domains \citep{wankhade2022survey}. While traditional sentiment analysis methods focus mainly on detecting sentiment at the sentence or document level, aspect-based sentiment analysis is a more fine-grained approach through which particular aspects expressed in a text are identified with their corresponding sentiment \cite{10.1109/TKDE.2022.3230975}. For a movie review, e.g., in this way not just the overall sentiment expressed by the review is identified but also, for example, whether the reviewer liked or disliked the score or camera work. Such more fine-grained insights are particularly valuable to businesses as they provide better insights into the needs of customers.

While datasets for traditional sentiment analysis are widely available (see e.g. \citet{app13074550, kenyon-dean-etal-2018-sentiment, 8474783}, and \citet{oro40660} for an overview of popular datasets), in part because they can be gathered in an automated fashion from services that use a combination of a score (e.g. in the form of stars) alongside with a textual review, the number of datasets available for aspect-based sentiment analysis is much more restricted (see e.g. \citet{8976252} and \citet{hua2024systematic}). Additionally, most of the existing datasets either contain a small number of aspects per item or all aspects in one item have the same polarity \citep{jiang-etal-2019-challenge}.

In this paper, we introduce a new, manually annotated, dataset for aspect-based sentiment analysis, that consists of 10,814 reviews for brick-and-mortar retail stores, scraped from Google Maps. The reviews cover different countries and languages and have been annotated with eight different aspect categories (see Table \ref{tab:aspects}) resulting in a total of 16,994 labels. The dataset is available on GitHub\footnote{\url{https://github.com/Responsible-NLP/ABSA-Retail-Corpus}}. A detailed datasheet \citep{gebru2021datasheets} for the corpus can be found in Appendix \ref{sec:datasheet}.

In addition, we present a Large Language Model (LLM)-based baseline for the newly introduced dataset comparing the performance of Meta's LLaMa-3 and OpenAI's GPT-4. The results show that while both models perform well with an accuracy of more than 85\%, GPT-4 consistently outperforms LLaMa-3 across aspects and metrics.

\begin{table*}
\centering
\caption{Sample of scraped data}
\label{tab:scraped_data}
\begin{tabular}{l l l l l}
\hline
\textbf{Country} & \textbf{City} & \textbf{Published At} & \textbf{Text} & \textbf{Stars} \\
\hline
Belgium & Maasmechelen & 18-03-2023 & Najbolja i najkvalitetnija roba & 5 \\ 
France  & Serris        & 29-12-2019 & Great prices                     & 5 \\ 
Italy   & Marcianise    & 21-11-2023 & Ho scoperto questo negozio grazie... & 5 \\
Belgium & Mechelen      & 30-11-2017 & Mooie propere zaak maar verkoper... & 3 \\ \hline
\end{tabular}
\end{table*}

\section{Related Work}

Compared to ``traditional'' sentiment analysis, aspects-based sentiment analysis presents a more complex challenge, encompassing two distinct stages: identifying all aspects described, and subsequently determining the sentiment towards each aspect. 

\subsection{LLMs for Aspect-Based Sentiment Analysis}
As for most NLP applications, recent literature about aspect-based sentiment analysis has mainly focused on the performance of LLMs. Recent studies that compared the performance of LLMs against smaller language models (SLMs), like BERT, across a spectrum of sentiment analysis tasks, including conventional sentiment classification and aspect-based analysis, found that LLMs exhibit proficiency in simpler tasks, such as sentiment classification, but encounter difficulties in tasks requiring nuanced understanding or structured sentiment information \citep{zhang-etal-2024-sentiment, machalikova2023utilizing, han2023design}.

In few-shot learning scenarios, however, where annotation resources are limited, LLMs have shown superior performance \citep{zhang-etal-2024-sentiment}. According to \citet{10.1007/978-3-031-51940-6_7}, LLMs also outperform smaller models in such tasks as predicting ratings of businesses based on online reviews, and leveraging aspect-based sentiment analysis techniques. Moreover, LLMs have introduced innovative methodologies for context-aware analysis, as shown by \citet{jeong2024aspect} in the context of hotel complaint reviews. 

Comparative analyses between GPT-3.5, BERT, RoBERTa, and LLaMA report superior performance of GPT-3.5, specifically in predicting product review ratings post fine-tuning \citep{roumeliotis2024llms}. 

\citet{krugmann2024sentiment} report that using a zero-shot nature, LLMs can not only compete with but in some cases also surpass traditional transfer learning methods in terms of sentiment classification accuracy. Additionally, studies emphasize the competitive performance of GPT-3.5 model in discerning nuanced sentiments like irony within social media tweets, achieved through \textit{prompt engineering} without explicit training \citep{carneros2023comparative} Moreover, effective prompting engineering and \textit{fine-tuning} are identified as crucial factors for achieving enhanced outputs and cost efficiency, further accentuating the potential of LLMs in customer satisfaction analysis and industry practices \citep{roumeliotis2024llms}.

Comparative studies between LLMs and lexicon-based methods, show that LLMs clearly outperform such methods, while being particularly good in annotating sentiment analysis data and achieving over 94\% accuracy in long-form sentiment reviews from Twitter social media users and Amazon customers, owing to its prowess in handling emojis, sarcasm, and contextual nuances \citep{belal2023leveraging}. Notably, the literature suggests that GPT's integration into business customer sentiment analysis reveals its potential to significantly enhance understanding of customer sentiments, offering valuable insights for decision-making processes by comprehending both general sentiments and nuanced factors within customer texts \citep{sudirjo2023application}.

\subsection{Datasets for Aspect-Based Sentiment Analysis}

Table \ref{tab:existingcorpora} shows a list of existing datasets for aspect-based sentiment analysis. While a number of datasets exists, the vast majority of them only covers the English language. Multilingual data sets, such as the one introduced in this paper, are particularly rare. With more than 10,000 annotated instances, the dataset introduced in this article is also one of the largest data sets for aspect-based sentiment analysis that is currently available.

\begin{table*}
    \centering
    \begin{tabular}{l l l l l}
    \hline
        \textbf{Dataset} & \textbf{Domain} & \textbf{Lang.} & \textbf{Size} & \textbf{Sources} \\\hline
        SemEval 2014 & Service and Product Reviews & English & 7,686 & \citet{pontiki-etal-2014-semeval} \\
        SemEval 2015 & Service and Product Reviews & English & 5,596 & \citet{pontiki-etal-2015-semeval}\\
        SemEval 2016 & Service and Product Reviews & 8 & 6,243 & \citet{pontiki-etal-2016-semeval}\\
        Pars-ABSA & Service Reviews & Persian & 5,602 & \citet{shangipour-ataei-etal-2022-pars}\\ 
        Foursqaure & Service Reviews & 	English & 585 & \citet{brun-nikoulina-2018-aspect}\\ 
        ACOS &  Service and Product Reviews & 	English & 6,362 & \citet{cai-etal-2021-aspect}\\ 
        SentiHood & Neighbourhood Q\&A	& English &	5,215 & \citet{saeidi-etal-2016-sentihood}\\
        MAMS & Service Reviews & English & 13,854 & \citet{jiang-etal-2019-challenge}\\
       \textit{Our dataset} & Service Reviews & 45 & 10,814 &\\\hline
    \end{tabular}
    \caption{Existing datasets for aspect-based sentiment analysis}
    \label{tab:existingcorpora}
\end{table*}

\section{Corpus}
\label{sec:corpus}
The data collection for the corpus relied on Google Maps 
reviews due to their accessibility and richness. While other publicly available datasets, such as those from product reviews or social media platforms, exist, many lack the level of granularity necessary for aspect-based sentiment analysis. These datasets typically emphasize general sentiment or aggregate ratings, which limits their suitability for examining specific aspects of customer feedback. Google Maps represent a robust platform where customers can leave reviews about a specific store or location. With its vast user base and widespread popularity, Google Maps is one of the most prominent platforms for user-generated reviews.

To efficiently scrape the reviews from Google Maps, the service Apify\footnote{\url{https://apify.com/compass/google-maps-reviews-scraper}} was utilized. Privacy and personal data were primary concerns during the data collection process. Despite Google Maps reviews being publicly accessible, it was important to be cautious to ensure compliance with privacy standards. Apify's functionality enabled the selection of only the essential columns, omitting personal identifiers such as the name/nickname of the reviewer. Consequently, the collected data only included the review text, star rating, timestamp, and the location of the store (country and city). A small sample of such can be seen in Table \ref{tab:scraped_data}.

\subsection{Data Cleaning and Augmentation}
A total of 24,361 reviews were collected in that way. Many reviews contained only a star rating without any textual review, these entries were excluded from the dataset, resulting in a final dataset of 10,814 reviews. Since the dataset contains reviews in a variety of languages, an additional column was added to encode the language of each review. The Google Translate API, accessed via the Python library \texttt{googletrans}, was utilized to detect the language of the reviews. The languages are encoded in ISO-639 format. The dataset also contains entries with unidentified languages and entries that contain only symbols or emojis. Additionally, the publication time, which was in textual format in the raw data, was converted to ISO-8601 format.

\subsection{Data Annotation}

The most important decision that had to be made before the data annotation is the definition of aspect categories. \textit{Aspect categories} are higher-level concepts that pool different \textit{aspect terms} to allow for more structured insights \citep{hua2024systematic}. The selection of the aspect categories was based on existing literature (particularly \citet{kang2022study, ramaswami2003reading, fakhira2023content}) and interviews with domain experts, to ensure that the chosen categories are relevant from both a scientific and a practitioner perspective. The eight aspect categories that have been derived from this process are shown in Table \ref{tab:aspects}. They cover a broad range of customer experiences and provide valuable insights into different facets of the reviews and the businesses behind them. Each of these categories, if identified in a review, was assigned a sentiment label (negative, positive, or neutral). 

The dataset was manually labeled by the authors. Manual labeling, though time-consuming, is critical for ensuring high-quality data annotations. Given the labor-intensive nature of this task, a custom labeling tool was developed to facilitate the process. Similar approaches have been employed in other studies. For instance, the SemEval-2014 task 4 involved creating annotation guidelines and tools for manual labeling to build benchmark datasets \citep{pontiki-etal-2014-semeval}. \citet{do2020automated} surveyed various tools and methods developed to assist in aspect-level sentiment annotation, while \citet{li2012generic} presented a tool designed to create high-quality training data through manual annotation.

\begin{figure}
    \centering
    \includegraphics[width=\linewidth]{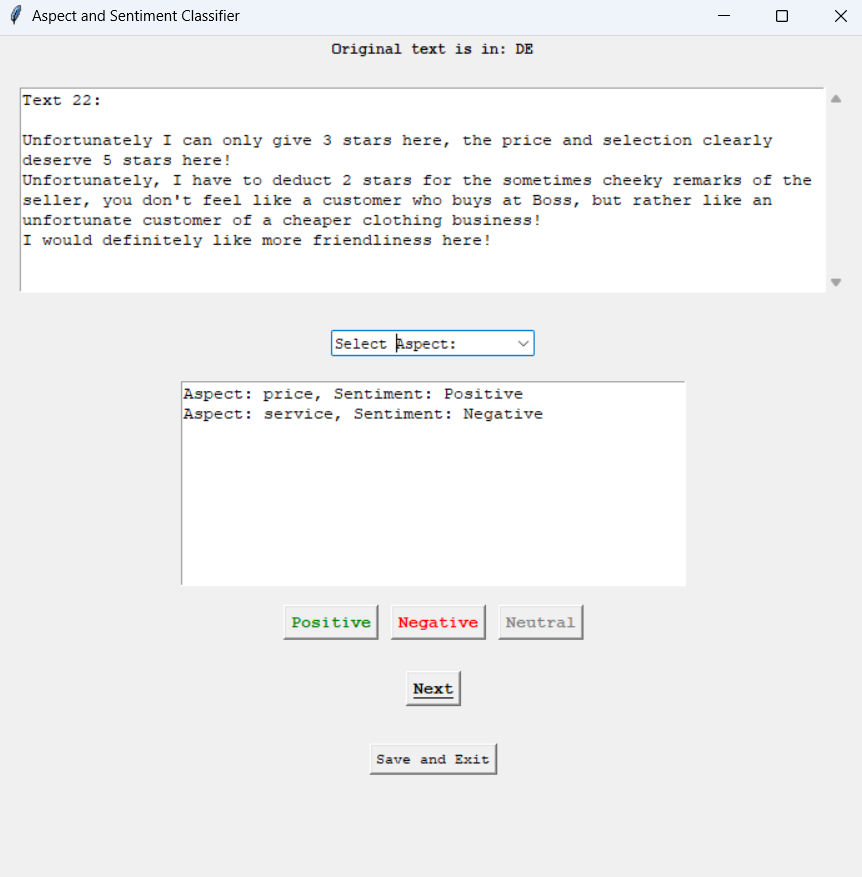}
    \caption{Labeling tool}
    \label{fig:tool}
\end{figure}

Figure \ref{fig:tool} shows the tool that was developed to annotate the dataset. It allowed the authors to review the text, select the relevant aspect, and assign the corresponding sentiment. The tool features a user-friendly interface with functionalities such as language detection and translation to English, sentiment selection, and annotation saving.

10\% of the data was annotated by two annotators to conduct an inter-annotator agreement study. Using Krippendorff's Alpha, the inter-annotator agreement is $\alpha = 0.71$, a value that is comparable to other ABSA datasets and in general indicates a reliable annotation. An in-depth analysis revealed only two instances in which the annotators chose the same aspect but different polarities. In both cases it was not a direct contradiction, as in a positive and a negative label at the same time, but one label was neutral. An analysis on aspect-level shows that more specific aspects like price and service show higher agreement, while more general categories like store and ``general'' show lower agreement. A reoccurring pattern is that often, annotators agree on most aspects, but one annotator adds a single additional aspect (like general), thereby decreasing agreement.

\begin{table}
\caption{Chosen aspect categories and their description}
\begin{tabular}{l p{5.5cm}}
\hline
\textbf{Aspect} & \textbf{Explanation} \\
\hline
\textit{Product} & Encompasses clothing collection, item quality, variety, display, and selection. \\
\textit{Service} & Includes staff, assistance, crew, employee attitude, handling, and hospitality. \\
\textit{Brand} & Pertains to overall brand perception. \\
\textit{Price} & Relates to the cost of products and services, including promotions and discounts. \\
\textit{Store} & Covers specific shop location, atmosphere, and environment. \\
\textit{Online} & Concerns the online ordering experience. \\
\textit{Return} & Includes the experience of returning an item for both physical or online procurement. \\
\textit{General} & Overall experience of the customer without a specific aspect mention. \\
\hline
\end{tabular}
\label{tab:aspects}
\end{table}

\subsection{Data Analysis}

Out of the 10,814 reviews in the dataset, 4,838 (or 44.7\%) contain more than one aspect. On average, each review contains 1.6 aspects. Figure \ref{fig:aspectCnt} shows how often each aspect category occurs in the dataset. The most frequently occurring aspect category is service, which is mentioned in 6,065 reviews. The least mentioned category, online, only occurred in 51 reviews, which is not surprising, given that the reviews were specifically collected for physical store locations.

The majority of the aspects mentioned in the dataset are positively connoted, as shown in Figure \ref{fig:sentiment_share}. Only for the aspect category return, the majority of the reviews expresses a negative sentiment. The most balanced aspect category is online, in which 55\% of all mentioned aspects are positive, 8\% neutral, and 37\% negative. Neutral aspects are rarely mentioned across all categories. The highest share of neutral aspects can be found for the category general, with a little over 8\%.

\begin{figure}
     \centering
     \begin{tikzpicture}
   \begin{axis}[
   width  = \linewidth,
   xbar,
   height = 25cm,
   flexible yticklabels from table={data/lang.dat}{lang}{col sep=space},
     x axis line style = { opacity = 0 },
     tickwidth         = 0pt,
     enlarge y limits={value=0.01,auto},
     yticklabel style={text height=1.5ex}, 
     ytick=data,
     xmajorgrids = true,
   ]
  
   \addplot [blue,fill=blue!70!white,draw=none] table [x=cnt, y expr=\coordindex]{data/lang.dat};

   \end{axis}
 \end{tikzpicture}
   \caption{Number of reviews per language (ISO-639)}
   \label{fig:langCnt}
 \end{figure}
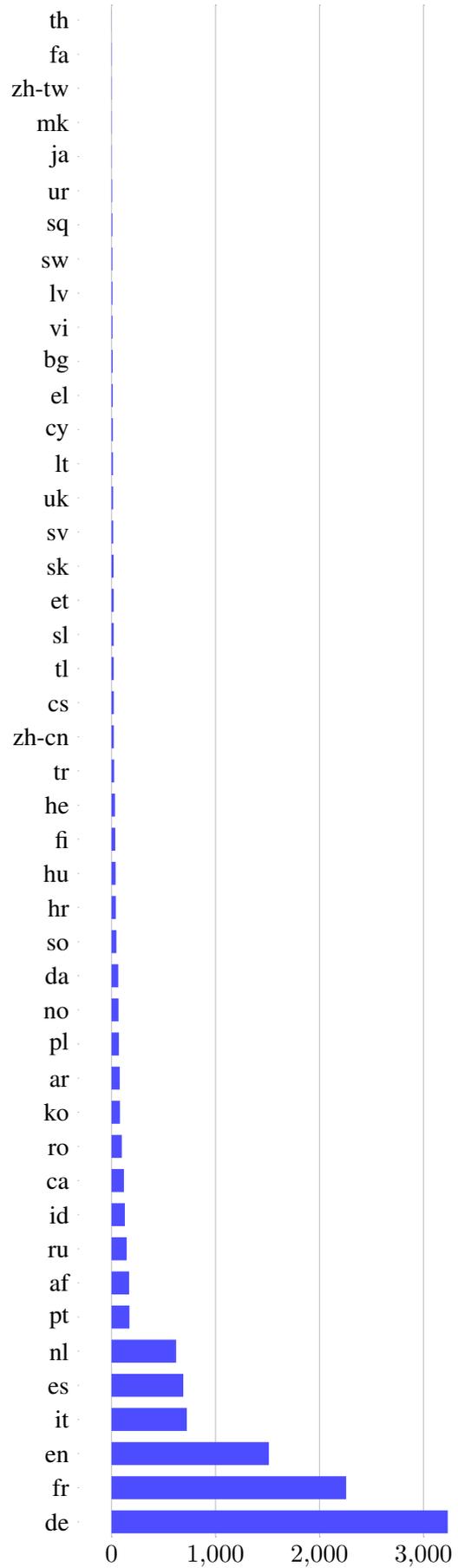

The reviews in the dataset have an average length of 121 characters, ranging from just one character (mostly emojis) to 3,735 characters. They cover stores from nine different European countries (Germany, France, Netherlands, Italy, Spain, Austria, Belgium, Portugal, and Switzerland; in descending frequency of occurrence) and are written in 45 different languages (see Figure \ref{fig:langCnt} for a distribution of the languages).

\begin{figure}

    \centering
    \begin{tikzpicture}
  \begin{axis}[
  width  = \linewidth,
  ybar,
  height = 4cm,
    x axis line style = { opacity = 0 },
    tickwidth         = 0pt,
    enlarge x limits=0.05,
    xticklabel style={rotate=45},
    symbolic x coords = {store, service, product, return, brand, price, online, general},
    xtick = data,
    ymin = 0,
    ymajorgrids = true,
  ]
  
  \addplot[blue,fill=blue!70!white,draw=none] coordinates{ (store, 3064) (service, 6065) (product, 2958) (return, 113) (brand, 894) (price, 1994) (online, 51) (general, 1855)};

  \end{axis}
\end{tikzpicture}
  \caption{Occurrences of each aspect category}
  \label{fig:aspectCnt}

    \begin{tikzpicture}
        \begin{axis}[
        width  = \linewidth,
            ybar stacked,
        	bar width=15pt,
            yticklabel=\pgfmathprintnumber{\tick}\,$\%$,
            enlargelimits=0.15,
            legend style={at={(0.5,-0.30)},
              anchor=north,legend columns=-1},    
            symbolic x coords = {store, service, product, return, brand, price, online, general},
            xtick=data,
            x tick label style={rotate=45},
            x axis line style = { opacity = 0 },
            ymajorgrids = true,
            tickwidth         = 0pt,
            ]
        
        
        \addplot+[ybar, green!60, pattern color=white, postaction={pattern=north east lines}] table [x=aspect, y=pos,col sep=comma]{data/share_sentiment.csv};
        
        \addplot+[ybar, gray!60] table [x=aspect, y=neu,col sep=comma]{data/share_sentiment.csv};
        
        \addplot+[ybar, red!60, pattern color=white, postaction={pattern=north west lines}] table [x=aspect, y=neg,col sep=comma]{data/share_sentiment.csv};
            
        \legend{\strut positive, \strut neutral, \strut negative}
        \end{axis}
    \end{tikzpicture}

\caption{Share of positive, neutral, and negative aspects per category\label{fig:sentiment_share}}
\end{figure}

\section{Experimental Set-Up}

To establish a baseline in aspect-based sentiment analysis for the newly introduced dataset, we conducted an experiment comparing the performance of the open weights model LLaMa-3 \citep{dubey2024llama} and the proprietary GPT-4 model \citep{achiam2023gpt}.

LLaMa-3 (\texttt{Meta-Llama-3-70B-Instruct}) was integrated through the HuggingFace library. To enhance performance and efficiency, quantization techniques were applied, reducing the model’s weight precision via \texttt{BitsAndBytesConfig}, which optimized computational resources, particularly for local deployment on GPU clusters.

GPT-4 was used through Azure's OpenAI Service. The implementation leveraged LangChain and its \texttt{PromptTemplate} component. 

For both models, we used prompt engineering \citep{brown2020language, radford2019language, gao2021making, lu2021fantastically}, and both system and user prompts, to facilitate the aspect-based sentiment analysis. \textit{System prompts} are designed to define the role of the language model and establish operational guidelines to ensure consistency in responses. For both GPT-4 and LLaMA-3, the system prompt instructed the model to perform aspect-based sentiment analysis for a specific company. This approach is known as intent classification. The system prompt was implemented differently for each model, reflecting their respective frameworks. GPT-4 used LangChain’s \texttt{SystemMessage} object to deliver the system instructions, while LLaMA-3 structured the system message as part of its chat template.

Despite syntactic differences, both implementations enforce a structured response format by defining system behavior upfront. LLaMA-3 specifies message roles such as \textit{"system"} and \textit{"user"} within its message list (see Listing \ref{lst:llama3system}), while GPT-4 utilizes LangChain’s \textit{langchain\_core.messages} framework to differentiate between system and human messages (see Listing \ref{lst:gpt4system}). These system prompts establish a consistent operational framework, ensuring the model generates precise and task-specific sentiment analysis responses.

User prompts provide the actual reviews and define the task parameters, guiding the model in identifying relevant aspects and classifying sentiments. The construction of these prompts is essential for ensuring accurate analysis, as they help the model distinguish between various aspects of a review and interpret sentiments effectively. To enhance performance, the prompts can incorporate structured instructions, few-shot examples, delimiters, and attention mechanisms.

\begin{table*}[ht]
    \caption{Precision, Recall, F1 Score, and Accuracy of the aspect-based sentiment analysis per model and aspect}
    \centering
    \begin{tabular}{ l r r r r r r r r}
    \hline    
     & \multicolumn{4}{c}{\textbf{GPT-4}} & \multicolumn{4}{c}{\textbf{LLaMA-3}}\\
    \textbf{Aspect} & \textbf{Prec.} & \textbf{Recall} & \textbf{F1} & \textbf{Accur.} & \textbf{Prec.} & \textbf{Recall} & \textbf{F1} & \textbf{Accur.} \\ \hline
    Store & \textbf{73.00\%} & 73.61\% & \textbf{73.31\%} & 71.23\% & 59.02\% & \textbf{80.13\%} & 67.97\% & \textbf{75.53\%}\\ 
    Service & \textbf{93.88\%} & \textbf{98.00\%} & \textbf{95.89\%} & \textbf{95.31\%} & 93.43\% & 96.44\% & 94.91\% & 94.05\%\\ 
    Product & \textbf{65.34\%} & 92.98\% & \textbf{76.75\%} & \textbf{88.39\%} & 58.12\% & \textbf{93.70\%} & 71.74\% & 87.82\%\\ 
    Return & 55.13\% & \textbf{77.48\%} & \textbf{64.42\%} & \textbf{76.11\%} & \textbf{62.89\%} & 54.46\% & 58.37\% & 53.98\%\\ 
    Brand & \textbf{57.92\%} & 70.65\% & \textbf{63.66\%} & 65.66\% & 39.81\% & \textbf{84.92\%} & 54.21\% & \textbf{80.43\%}\\ 
    Price & \textbf{82.71\%} & \textbf{85.86\%} & \textbf{84.26\%} & \textbf{79.23\%} & 79.04\% & 80.84\% & 79.93\% & 73.09\%\\ 
    Online & 29.10\% & \textbf{82.98\%} & \textbf{43.09\%} & \textbf{76.47\%} & \textbf{30.09\%} & 72.34\% & 42.50\% & 66.67\% \\ 
    General & \textbf{49.75\%} & \textbf{78.09\%} & \textbf{60.78\%} & \textbf{73.92\%} & 43.96\% & 65.13\% & 52.49\% & 62.13\%\\ \hline
    \textbf{Micro avg.} & \textbf{74.48\%} & \textbf{87.58\%} & \textbf{80.50\%} & \textbf{83.80\%} & 66.84\% & 86.88\% & 75.55\% & 82.57\%\\ 
    \textbf{Macro avg.} & \textbf{63.35\%} &	\textbf{82.46\%} &\textbf{ 70.27\%} &	\textbf{78.29\%} & 58.30\% &	78.49\%	& 65.27\%	& 74.21\%\\\hline

    \end{tabular}
    \label{tab:resultsperaspect}
\end{table*}

One effective prompt engineering strategy is task decomposition, where a complex task is divided into smaller, more manageable steps. For instance:

\begin{displayquote} 
"\underline{First,} identify the aspects in the provided review from the given list, \underline{and then} find the customer sentiment (positive, neutral, or negative) for each of the aspects. ..." 
\end{displayquote}

Another key technique is few-shot prompting, where examples of correct outputs are included in the prompt to guide the model’s response.

\begin{displayquote} 
" ... You can follow the examples below:

Review: The product quality is great but the customer service is terrible.\ Aspect: product\ Sentiment: positive\ Aspect: service\ Sentiment: negative

Review: I love the location of the store. The collection and selection look great, however, the prices are too high.\ Aspect: store\ Sentiment: positive\ Aspect: product\ Sentiment: positive\ Aspect: prices\ Sentiment: negative

Review: All top, everything as I expected, recommend.\ Aspect: general\ Sentiment: positive ... " 
\end{displayquote}

Finally, attention mechanisms can be influenced by placing key instructions at the beginning and end of the prompt. For example:

\begin{displayquote} 
"First, identify the aspects in the provided review from the given list, and then find the customer sentiment (positive, neutral, or negative) for each aspect.

Make sure to take into account the difference in language, cultural aspects, sarcasm, emojis, and other linguistic behaviors when interpreting and assessing the reviews.\ ...

...\ Remember to strictly focus only on the aspects from the list and reply only with the answer in the following JSON format:

[{{"aspect1": "sentiment"}},\ ...\ {{"aspectN": "sentiment"}}]"\ 
\end{displayquote}

By placing important contextual elements at the beginning and specifying output format at the end, the model is guided to prioritize crucial information while maintaining structured responses. The final prompts used in the experiment can be found in Appendix \ref{sec:userprompts}.

\section{Results}
Table \ref{tab:resultsperaspect} shows the evaluation of the aspect-based sentiment analysis experiment. Overall, GPT-4 outperformed LLaMA-3 in every single metric, with the widest gap in precision and the narrowest gap in recall. Given the imbalance between positive and negative reviews, the difference in accuracy of both models is only about 1.2 percentage points, despite the larger difference in precision. Only for the aspect categories ``store'' and ``brand'', LLaMa-3 outperformed GPT-4 in respect to accuracy. 

In high-frequency aspects such as ``service'' and ``product'', both models showed strong performance, with GPT-4 recording accuracies of 95.31\% and 88.39\%, respectively, compared to LLaMA-3’s 93.05\% and 87.82\%. Notably, LLaMA-3 demonstrated improvements in the ``product'' category, narrowing the gap with GPT-4. These results indicate that both models reliably handle frequently mentioned aspects, although GPT-4 retains a slight edge in overall robustness.

When addressing lower-frequency or more nuanced aspects such as ``return'', ``brand'', and ``online'', both models continued to face challenges, albeit with notable differences. GPT-4 demonstrated better performance in the ``return'' category, achieving an accuracy of 76.11\% compared to LLaMA-3’s 53.98\%. Similarly, in ``online'', GPT-4 outperformed LLaMA-3, recording accuracies of 76.47\% versus 66.67\%. These findings underscore GPT-4's greater capability to handle complex sentiment categories, though significant gaps remain. Both models struggled with the ``online'' aspect in terms of precision, with GPT-4 achieving 29.10\% and LLaMA-3 slightly higher at 30.09\%. These metrics highlight a broader limitation in capturing context-dependent nuances in less straightforward sentiment categories. In order for an aspect-based sentiment to be classified as correct, both the aspect and the sentiment expressed with it have to be extracted correctly. Notably, a large share of errors already occurs during the identification of aspects (see Table \ref{tab:aspectidentification}). The number of aspects that have been identified correctly but the sentiment was misclassified is relatively small.

\section{Error Analysis}
A deeper analysis of the errors made by both models revealed that LLaMA-3 exhibited a tendency to incorrectly identify aspects that are not present in the text based on mentioned keywords. For instance, in reviews such as ``Great store for its amazing service and help from the assistants'', LLaMA-3 frequently identified ``store'' as an aspect, whereas the correct interpretation according to the aspect categories defined in Table \ref{tab:aspects} would be ``service''. This over-sensitivity to mentions of keywords is also visible in Table \ref{tab:resultsperaspect}, where the precision for the aspect store is particularly low for LLaMA-3. Yet, given the prevalence of the aspect across the dataset, this over-sensitivity might also partially explain why this is one of just two aspect categories in which LLaMA-3 achieved a higher recall and accuracy than GPT-4.

Both models encountered difficulties in consistently handling aspects from the categories ``general'' and ``brand''. Reviews with broad or ambiguous sentiments about a brand in general, such as "Brand for Bosses," or "I love BOSS" for the clothing brand ``BOSS'' posed a significant challenge for both models. These reviews often led to inconsistent labeling, with models sometimes assigning both ``brand'' and ``general'' aspects or failing to distinguish between them altogether. Given the ambiguity of such statements, even human annotators would likely struggle to reach consensus, making this a particularly challenging area for automated analysis and labeling.

LLaMA-3 showed a tendency to classify reviews under the apsect category ``brand'' more frequently than GPT-4, which contributed to its lower accuracy for the aspect category ``general'', scoring 66.74\% comparing to 79.27\% for GPT-4. Simultaneously, LLaMa-3 outperformed GPT-4 in the aspect category ``brand'', being one of only two aspect categories where it outperformed GPT-4 model. This behavior, while indicative of an attempt to capture a broader range of sentiment, often led to increased misclassification rates when the sentiment was intended to be more general. The higher labeling frequency for ``brand'' by LLaMA-3 also aligns with its overall lower precision and F1 scores in this aspect, reflecting a trade-off between recall and precision.

\begin{table}
\caption{Aspect identification accuracy}
\centering
\begin{tabular}{l r r}
\hline
\textbf{Aspect} & \textbf{GPT-4} & \textbf{LLaMA-3} \\ \hline
Store & 74.47\% & \textbf{81.26\%} \\ 
Service & \textbf{98.05\%} & 96.57\% \\ 
Product & 93.33\% & \textbf{94.10\%} \\ 
Return & \textbf{77.88\%} & 54.87\% \\ 
Brand & 72.73\% & \textbf{85.71\%} \\ 
Price & \textbf{86.95\%} & 82.69\% \\ 
Online & \textbf{84.31\%} & 74.51\% \\ 
General & \textbf{79.27\%} & 66.74\% \\\hline
\textbf{Micro avg.} & \textbf{88.12\%} & 87.57\%\\ 
\textbf{Macro avg.} & \textbf{83.37\%} & 79.44\%\\\hline
\end{tabular}
\label{tab:aspectidentification}
\end{table}

\section{Conclusion}

This paper introduced a new multilingual corpus for aspect-based sentiment analysis that is based on more than 10,000 reviews of brick-and-mortar stores and was manually labeled with eight aspect categories, namely product, service, brand, price, store, online, return, and general (see Table \ref{tab:aspects}).

Additionaly, an experiment was conducted to establish a baseline for LLM-based aspect-based sentiment analysis on the newly introduced corpus by comparing the performance of GPT-4 and LLaMA-3. The results indicate that both models proficiently identify elements and attitudes in customer reviews, with GPT-4 continuously surpassing LLaMA-3 in precision, recall, and accuracy. Although both models achieved accuracy exceeding 85\%, they performed insufficiently for the ``store'' and ``brand'' aspects, indicating areas for enhancement. 
From an application perspective, in addition to the performance, it is also worth considering the potential costs of different approaches and models. At the time of conducting the experiments, GPT-4 was queried through the Azure OpenAI API at a total cost of \$240.60 for the complete dataset, or an average cost of \$0.022 per review. With the price for one million input tokens being around \$2.50 and the price for one million output tokens being around \$10.00. LLaMA-3, on the other hand, cannot just be self hosted, but also used through cloud providers like groq who charge in the realm of \$0.59 / \$0.79 per one million input / output tokens, providing significantly cheaper options.

\section*{Ethics}

By using public reviews and ensuring during both collection and annotation of the data that no identifiable information is contained in the reviews, we tried to ensure that our work has no direct adverse effects to anyone. Nevertheless, given the nature of the task, companies could use an approach like the one outlined in this work to try to automatically assess job performance of workers in physical store locations (e.g. by focusing on the aspect category service). However, we believe that in practice, the danger for such applications is relatively low given that most reviews (and none in our dataset) name specific employees or provide other information that could be used for the identification of individual employees like exact data and time of an interaction. While the environmental impact of LLMs is mostly discussed with regard to their training and fine-tuning, ever larger models also have an increasingly significant environmental impact during inference, which also holds true for this work.
 
\section*{Limitations}

This study faced limitations inherent to the rapidly evolving field of LLM research. The introduction of newer models may render some aspects of this study's model selection less timely, although its fundamental methodologies remain relevant. Computational limitations also restricted fine-tuning and advanced quick engineering for LLaMA-3, potentially affecting its performance. Furthermore, the hand annotated dataset, albeit comprehensive, introduced subjectivity in aspect and sentiment classification, with ambiguous phrases and linguistic variances presenting hurdles to consistency. Lastly, the predefined aspect categories may not generalize across all use cases, making prompt design sensitive to initial definitions.


\bibliography{custom}
\newpage
\appendix

\section{Prompts}
\label{sec:systemprompts}
\label{sec:userprompts}
\begin{lstlisting}[language=Python, caption={LLaMA-3 system prompt},label={lst:llama3system}]
messages = [ {"role": "system", "content": """You are a helpful assistant that performs aspect based sentiment analysis for Hugo Boss! Do not communicate back, just provide the answer in the requested format"""}, {"role": "user", "content": text}, ]
prompt = output.tokenizer.apply_chat_template( messages, tokenize=False, add_generation_prompt=True ) 
\end{lstlisting}

\begin{lstlisting}[language=Python,caption={GPT-4 system prompt},label={lst:gpt4system}]
system_prompt = """You are a helpful assistant that performs aspect based sentiment analysis for Hugo Boss! Do not communicate back, just provide the answer in the requested format."""

prompt_value = StringPromptValue(text=chat_prompt_with_values)

output = llm.invoke([ SystemMessage(content=system_prompt), HumanMessage(content=prompt_value.text), ])
\end{lstlisting}


\begin{lstlisting}[language=Python, caption={LLaMA-3 user prompt},label={lst:llama3user}]
text = f"""First, identify the aspects in the provided review from the given list, and then find the customer sentiment (positive, neutral, or negative) for each of the aspect.
Make sure to take into account the difference in the language, cultural aspects, sarcasm, emojis, and other linguistic behaviours when interpreting and assessing the reviews.

Aspects: [Product (collection, item, quality, variety, display, selection), Service (staff, assistance, crew, employee, attitude, handling, hospitality), Brand, Price, Store (shop, location, atmosphere), Online (order), Purchase, Return, General (overall shopping experience)]

You can follow the examples below:

Review: The product quality is great but the customer service is terrible.
Aspect: product
Sentiment: positive
Aspect: service
Sentiment: negative

Review: I love the location of the store. The collection and selection looks great, however the prices are too high. 
Aspect: store
Sentiment: positive
Aspect: product
Sentiment: positive
Aspect: prices
Sentiment: negative

Review: All top, everything as I expected, recommend.
Aspect: general
Sentiment: positive

Now proceed with the following review: ```{review}```

Remember to strictly focus only on the aspects from the list and reply only with the answer in the following JSON format:
[{{"aspect1": "sentiment"}},
...
{{"aspectN": "sentiment"}}]
"""
\end{lstlisting}

\begin{lstlisting}[language=Python, caption={GPT-4 user prompt},label={lst:gpt4user}]
First, identify the aspects in the provided review from the given list, and then find the customer sentiment (positive, neutral ot negative) for each of the aspect.

Aspects: [Product (collection, item, quality, variety, display), Service (staff, assistance, crew, employee, attitude, handling, hospitality), Brand, Price, Store (shop, location, atmosphere), Online (order), Return, General (overall shopping experience)]

You can follow the examples below:
review: "The product quality is great but the customer service is terrible."
analysis: "[{{"product": "positive", "service": "negative"}}]"

review: "I love the location of the store. The collection looks great, however the prices are too high.",
analysis: "[{{"store": "positive", "product": "positive", "price": "negative"}}]"

review: "All top, everything as I expected, recommend.",
analysis: "[{{"general": "positive"}}]"

Make sure to take in account the difference in the language, cultural aspects, sarcasm, emojis and other linguistic behaviours when interpreting and assessing the reviews. Remember to strictly reply only with the answer in the following JSON format:
[{{"aspect1": "sentiment"}},
...
{{"aspectN": "sentiment"}}]
\end{lstlisting}

\section{Datasheet}
\label{sec:datasheet}

\subsection{Motivation for Dataset Creation}

\textcolor{blue}{\textbf{Why was the dataset created?} (e.g., were there specific
tasks in mind, or a specific gap that needed to be filled?)}

The dataset was created to enable aspect-based sentiment analysis on customer reviews using Large Language Models (LLMs).

\textcolor{blue}{\textbf{What (other) tasks could the dataset be used for?} Are
there obvious tasks for which it should not be used?}

The dataset could also be used for traditional sentiment analysis given it also contains star ratings for each review. 

\textcolor{blue}{\textbf{Has the dataset been used for any tasks already?} If so,
where are the results so others can compare (e.g., links to
published papers)?}

This paper is the first to use the dataset.

\textcolor{blue}{\textbf{Who funded the creation of the dataset?} If there is an
associated grant, provide the grant number.}

The data collection was supported by the consulting firm Metyis (\url{https://metyis.com/}).

\subsection{Dataset Composition}

\textcolor{blue}{\textbf{What are the instances?} (that is, examples; e.g., documents, images, people, countries) Are there multiple types of instances? (e.g., movies, users, ratings; people, interactions between them; nodes, edges)}

Each instance consists of a customer review for a brick-and-mortar store, scarped from Google maps.

\textcolor{blue}{\textbf{Are relationships between instances made explicit in
the data (e.g., social network links, user/movie ratings, etc.)?}}

No.

\textcolor{blue}{\textbf{How many instances of each type are there?}}

The dataset consists of 10,814 reviews.

\textcolor{blue}{\textbf{What data does each instance consist of?} “Raw” data
(e.g., unprocessed text or images)? Features/attributes? Is there a label/target associated with instances? If the
instances are related to people, are subpopulations identified
(e.g., by age, gender, etc.) and what is their distribution?}

In addition to the country and city of the store that is reviewed, the date the review was published at, its text, and the star rating are part of each instance. The language of the review is automatically annotated, while aspects and their sentiments have been manually annotated.

\textcolor{blue}{\textbf{Is everything included or does the data rely on external
resources?} (e.g., websites, tweets, datasets) If external
resources, a) are there guarantees that they will exist, and
remain constant, over time; b) is there an official archival
version. Are there licenses, fees or rights associated with
any of the data?}

Everything is included in the dataset.

\textcolor{blue}{\textbf{Are there recommended data splits or evaluation measures?} (e.g., training, development, testing; accuracy/AUC)}

Since the dataset is designed for zero-shot classification, there is no recommended split. Given the imbalanced distribution of positive and negative sentiments, we recommend and evaluation measure that takes this into account, like F1-score.

\textcolor{blue}{\textbf{What experiments were initially run on this dataset?}
Have a summary of those results and, if available, provide
the link to a paper with more information here.}

The dataset was initially used for the evaluation of the aspect-based sentiment analysis capabilities of LLMs.

\subsection{Data Collection Process}

\textcolor{blue}{\textbf{How was the data collected?} (e.g., hardware apparatus/sensor, manual human curation, software program, software interface/API; how were these constructs/measures/methods validated?)}

The data was collected using Apify (\url{https://apify.com/compass/google-maps-reviews-scraper}).

\textcolor{blue}{\textbf{Who was involved in the data collection process?} (e.g.,
students, crowdworkers) How were they compensated? (e.g.,
how much were crowdworkers paid?)}

The data was collected by fully-qualified lawyers during their usual work-time. All participants worked for organizations that pay according to the collective labor agreement for public service workers in German states.

\textcolor{blue}{\textbf{Over what time-frame was the data collected?} Does the
collection time-frame match the creation time-frame?}

The data was collected in 2024. The reviews were written between 2012 and 2024.

\textcolor{blue}{\textbf{How was the data associated with each instance acquired?} Was the data directly observable (e.g., raw text,
movie ratings), reported by subjects (e.g., survey responses),
or indirectly inferred/derived from other data (e.g., part of
speech tags; model-based guesses for age or language)? If
the latter two, were they validated/verified and if so how?}

The reviews themselves and their metadata was directly observable, the language of the reviews was automatically derived using the Google Translate API, the aspects and their sentiments were manually annotated by the authors.

\textcolor{blue}{\textbf{Does the dataset contain all possible instances?} Or is
it, for instance, a sample (not necessarily random) from a
larger set of instances?}

No, the dataset does not claim completeness in any sense.

\textcolor{blue}{\textbf{If the dataset is a sample, then what is the population?}
What was the sampling strategy (e.g., deterministic, probabilistic with specific sampling probabilities)? Is the sample representative of the larger set (e.g., geographic coverage)?
If not, why not (e.g., to cover a more diverse range of instances)? How does this affect possible uses?}

The dataset spans multiple European countries and a time-frame of over a decade.

\textcolor{blue}{\textbf{Is there information missing from the dataset and why?}
(this does not include intentionally dropped instances; it
might include, e.g., redacted text, withheld documents) Is
this data missing because it was unavailable?}

Reviews that only consists of a star rating but do not provide any text have been excluded.

\subsection{Dataset Distribution}

\textcolor{blue}{\textbf{How is the dataset distributed?} (e.g., website, API, etc.;
does the data have a DOI; is it archived redundantly?)}

It is archived on GitHub (\url{https://github.com/Responsible-NLP/ABSA-Retail-Corpus}).

\textcolor{blue}{\textbf{When will the dataset be released/first distributed?} (Is
there a canonical paper/reference for this dataset?)}

Publication of the paper.

\textcolor{blue}{\textbf{What license (if any) is it distributed under?} Are there
any copyrights on the data?}

The annotations are licensed under CC-BY-SA 4.0.

\textcolor{blue}{\textbf{Are there any fees or access/export restrictions?}}

No.

\subsection{Dataset Maintenance}
\textcolor{blue}{\textbf{Who is supporting/hosting/maintaining the dataset?}
How does one contact the owner/curator/manager of the
dataset (e.g. email address, or other contact info)?}

See the GitHub repository.

\textcolor{blue}{\textbf{Will the dataset be updated?} How often and by whom?
How will updates/revisions be documented and communicated (e.g., mailing list, GitHub)? Is there an erratum?}

There are no plans to update the dataset unless important mistakes become clear.

\textcolor{blue}{\textbf{If the dataset becomes obsolete how will this be communicated?}}

On the GitHub page.

\textcolor{blue}{\textbf{Is there a repository to link to any/all papers/systems
that use this dataset?}}

Yes.

\textcolor{blue}{\textbf{If others want to extend/augment/build on this dataset,
is there a mechanism for them to do so?} If so, is there
a process for tracking/assessing the quality of those contributions. What is the process for communicating/distributing
these contributions to users?}

We would suggest to create a fork on GitHub.

\subsection{Legal \& Ethical Considerations}

\textcolor{blue}{\textbf{If the dataset relates to people (e.g., their attributes) or
was generated by people, were they informed about the
data collection?} (e.g., datasets that collect writing, photos,
interactions, transactions, etc.)}

There is no information about individuals in the data or was recorded during the annotation of the data.

\textcolor{blue}{\textbf{If it relates to other ethically protected subjects, have
appropriate obligations been met?} (e.g., medical data
might include information collected from animals)}

N.a.

\textcolor{blue}{\textbf{If it relates to people, were there any ethical review applications/reviews/approvals?} (e.g. Institutional Review
Board applications)}

N.a.

\textcolor{blue}{\textbf{If it relates to people, were they told what the dataset
would be used for and did they consent? What community norms exist for data collected from human communications?} If consent was obtained, how? Were the people
provided with any mechanism to revoke their consent in the
future or for certain uses?}

N.a.

\textcolor{blue}{\textbf{If it relates to people, could this dataset expose people
to harm or legal action?} (e.g., financial social or otherwise)
What was done to mitigate or reduce the potential for harm?}

N.a.

\textcolor{blue}{\textbf{If it relates to people, does it unfairly advantage or disadvantage a particular social group?} In what ways? How
was this mitigated?}

N.a.

\textcolor{blue}{\textbf{If it relates to people, were they provided with privacy
guarantees?} If so, what guarantees and how are these
ensured?}

N.a.

\textcolor{blue}{\textbf{Does the dataset comply with the EU General Data Protection Regulation (GDPR)?} Does it comply with any other
standards, such as the US Equal Employment Opportunity
Act?}

Yes, since only publicly available information was collected, the dataset complies with the GDPR and similar regulations.

\textcolor{blue}{\textbf{Does the dataset contain information that might be considered sensitive or confidential?} (e.g., personally identifying information)}

No.

\textcolor{blue}{\textbf{Does the dataset contain information that might be considered inappropriate or offensive?}}

No.

\end{document}